# Expert-Guided Subgroup Discovery:
# Methodology and Application

**Dragan Gamberger**                                    DRAGAN.GAMBERGER@IRB.HR
*Rudjer Bošković Institute, Bijenička 54*
*10000 Zagreb, Croatia*

**Nada Lavrač**                                         NADA.LAVRAC@IJS.SI
*Jožef Stefan Institute, Jamova 39*
*1000 Ljubljana, Slovenia*

## Abstract

This paper presents an approach to expert-guided subgroup discovery. The main step of the subgroup discovery process, the induction of subgroup descriptions, is performed by a heuristic beam search algorithm, using a novel parametrized definition of rule quality which is analyzed in detail. The other important steps of the proposed subgroup discovery process are the detection of statistically significant properties of selected subgroups and subgroup visualization: statistically significant properties are used to enrich the descriptions of induced subgroups, while the visualization shows subgroup properties in the form of distributions of the numbers of examples in the subgroups. The approach is illustrated by the results obtained for a medical problem of early detection of patient risk groups.

## 1. Introduction

This paper addresses the problem of subgroup discovery which can be defined as: given a population of individuals and a property of those individuals we are interested in, find population subgroups that are statistically 'most interesting', e.g., are as large as possible and have the most unusual statistical (distributional) characteristics with respect to the property of interest (Klösgen, 1996; Wrobel, 1997, 2001). Its main contribution is a new methodology supporting the process of expert-guided subgroup discovery. Specifically, we introduce a novel parametrized definition of rule quality used in a heuristic beam search algorithm, a rule subset selection algorithm incorporating example weights, the detection of statistically significant properties of selected subgroups, and a novel subgroup visualization method. An in-depth analysis of the proposed quality measure is provided as well. The proposed methodology has been applied to the medical problem of detecting and describing patient groups with high risk for artherosclerotic coronary heart disease (CHD).[1]

The paper organization is as follows. Algorithms for subgroup detection and selection, which are the main ingredients of the expert-guided subgroup discovery methodology, are described in Section 2. Section 3 presents: the coronary heart disease risk group detection problem, the discovered patient risk groups, their statistical characterization, visualization, medical interpretation and evaluation, including a discussion on the expert's role in the subgroup discovery process. Section 4 provides an in-depth analysis of the proposed rule quality measure for subgroup discovery including an experimental comparison with a selected cost-based quality measure. Finally, Section 5 provides links to the related work.

---

1. Algorithms for subgroup detection and selection have been implemented in the on-line Data Mining Server (Gamberger & Šmuc, 2001), publicly available at `http://dms.irb.hr` which can be tested in domains with up to 250 examples. A more sophisticated implementation of the algorithms is not available for public use.

## 2. Subgroup Discovery: Rule Induction and Selection

This section describes the two main steps of the overall subgroup discovery process: induction and selection of interesting subgroups. These two steps, as well as the whole descriptive induction process assume active expert involvement.

### 2.1 The Task of Expert-Guided Subgroup Discovery

The task of expert-guided subgroup discovery addressed in this work differs slightly from the subgroup discovery task defined in Section 1 and proposed by (Klösgen, 1996; Wrobel, 1997). Instead of defining an optimal measure for automated subgroup search and selection, here the goal is to *support* the expert in performing flexible and effective search of a broad range of optimal solutions. As a consequence, the decision of which subgroups will be selected to form the final solution is left to the expert. The task of the subgroup discovery algorithm is to enable the detection of rules describing potentially optimal subgroups, which are characterized by the property that they are correct for many target class cases (patients with coronary heart disease, in the example domain used in this work) and incorrect for all, or most of, non-target class cases (healthy subjects). Target class cases included into a subgroup are called *true positives* while non-target class cases incorrectly included into a subgroup are called *false positives*.

The particular expert-guided subgroup discovery task addressed in this work assumes the collaboration of the expert and the data analyst in repeatedly running a subgroup discovery algorithm with a goal of finding rules describing population subgroups which:

- have sufficiently large coverage,

- have a positive bias towards target class case coverage (have a sufficiently large true positive/false positive ratio)

- are sufficiently diverse for detecting most of the target population, and

- fulfill other experts' subjective measures of acceptability: understandability, simplicity and actionability.

In each iteration, the task of the subgroup discovery algorithm is to suggest one or more potentially optimal solutions. Section 2.2 describes a heuristic search algorithm SD, which can be used to construct many rules that are optimal with respect to an expert selected generalization parameter. Since many of the induced rules can be very similar, both in terms of their coverage and the selected features, the RSS algorithm described in Section 2.3 can be used to select a small number of distinct rules that are offered to the expert as potentially optimal solutions. Alternatively, subgroup discovery can be implemented within a 'weighted' covering algorithm DMS, as is the case in the publicly available Data Mining Server (Gamberger & Šmuc, 2001), which generates up to three best subgroups in every iteration.

## 2.2 The Subgroup Discovery Algorithm

The goal of the subgroup discovery algorithm SD, outlined in Figure 1, is to search for rules that maximize $q_g = \frac{TP}{FP+g}$, where $TP$ are true positives, $FP$ are false positives, and $g$ is a *generalization parameter*. High quality rules cover many target class examples and a low number of non-target examples. The number of tolerated non-target class cases, relative to the number of covered target class cases, is determined by parameter $g$. For low $g$ ($g \leq 1$), induced rules will have high specificity (low false alarm rate) since covering of every single non-target class example is made relatively very 'expensive'. On the other hand, by selecting a high $g$ value ($g > 10$ for small domains), more general rules will be generated, covering also non-target class instances.

**Algorithm SD: Subgroup Discovery**
**Input:**      $E = P \cup N$ ($E$ training set, $|E|$ training set size,
                 $P$ positive (target class) examples, $N$ negative (non-target class) examples)
                 $L$ set of all defined features (attribute values), $l \in L$
**Parameter:** $g$ (generalization parameter, $0.1 < g$, default value 1)
                 $min\_support$ (minimal support for rule acceptance)
                 $beam\_width$ (maximal number of rules in $Beam$ and $New\_Beam$)
**Output:** $S = \{TargetClass \leftarrow Cond\}$ (set of rules formed of $beam\_width$ best conditions $Cond$)
(1)      **for** all rules in $Beam$ and $New\_Beam$ ($i = 1$ to $beam\_width$) **do**
              initialize condition part of the rule to be empty, $Cond(i) \leftarrow \{\}$
              initialize rule quality, $q_g(i) \leftarrow 0$
(2)      **while** there are improvements in $Beam$ **do**
(3)          **for** all rules in $Beam$ ($i = 1$ to $beam\_width$) **do**
(4)              **for** all $l \in L$ **do**
(5)                  form a new rule by forming a new condition as a conjunction of the
                         condition from $Beam$ and feature $l$, $Cond(i) \leftarrow Cond(i) \wedge l$
(6)                  compute the quality of a new rule as $q_g = \frac{TP}{FP+g}$
(7)                  **if** $\frac{TP}{|E|} \geq min\_support$ **and if** $q_g$ is larger than any $q_g(i)$ in $New\_Beam$
                         **and if** the new rule is relevant **do**
(8)                      replace the worst rule in $New\_Beam$ with the new rule and
                             reorder the rules in $New\_Beam$ with respect to their quality
(9)              **end for** features
(10)         **end for** rules from $Beam$
(11)         $Beam \leftarrow New\_Beam$
(12)     **end while**

Figure 1: Heuristic beam search rule construction algorithm for subgroup discovery.

Varying the value of $g$ enables the expert to guide subgroup discovery in the $TP/FP$ space, in which $FP$ (plotted on the $X$-axis) needs to be minimized, and $TP$ (plotted on the $Y$-axis) needs to be maximized. The $TP/FP$ space is similar to the ROC (Receiver Operating Characteristic) space (Provost & Fawcett, 2001). The comparison of the ROC and $TP/FP$ space and the $g_q$ heuristic are analyzed in detail in Sections 2.4 and 4, respectively.

Algorithm SD takes as its input the complete training set $E$ and the feature set $L$, where features $l \in L$ are logical conditions constructed from attribute values describing the examples in $E$. For discrete (categorical) attributes, features have the form $Attribute = value$ or $Attribute \neq value$, for numerical attributes they have the form $Attribute > value$ or $Attribute < value$. To formalize feature construction, let values $v_{ix}$ ($x = 1..k_{ip}$) denote the $k_{ip}$ different values of attribute $A_i$ that appear in the positive examples and $w_{iy}$ ($y = 1..k_{in}$) the $k_{in}$ different values of $A_i$ appearing in the negative examples. A set of features $L$ is constructed as follows:

- For discrete attributes $A_i$, features of the form $A_i = v_{ix}$ and $A_i \neq w_{iy}$ are generated.

- For continuous attributes $A_i$, similar to Fayyad and Irani (1992), features of the form $A_i \leq (v_{ix} + w_{iy})/2$ are created for all neighboring value pairs $(v_{ix}, w_{iy})$, and features $A_i > (v_{ix} + w_{iy})/2$ for all neighboring pairs $(w_{iy}, v_{ix})$.

- For integer valued attributes $A_i$, features are generated as if $A_i$ were both discrete and continuous, resulting in features of four different forms: $A_i \leq (v_{ix} + w_{iy})/2$, $A_i > (v_{ix} + w_{iy})/2$, $A_i = v_{ix}$, and $A_i \neq w_{iy}$.

There is no theoretical upper value for the user-refined $g$ parameter, but in practice the suggested upper limit should not exceed the number of training examples. For instance, suggested $g$ values in the Data Mining Server are in the range between 0.1 and 100, for analysing data sets of up to 250 examples. The choice of $g$ should be adjusted both to the size of the data set and to the proportion of positive examples in the set.

Algorithm SD has two additional parameters which are typically not adjusted by the user. The first is $min\_support$ (default value is $\sqrt{P}/E$, where $P$ is the number of target class examples in $E$) which indirectly defines the minimal number of target class examples which must be covered by every subgroup. The second is $beam\_width$ (default value is 20) which defines the number of solutions kept in each iteration. The output of the algorithm is set $S$ of $beam\_width$ different rules with highest $q_g$ values. The rules have the form of conjunctions of features from $L$.

The algorithm initializes all the rules in $Beam$ and $New\_beam$ by empty rule conditions. Their quality values $q_g(i)$ are set to zero (step 1). Rule initialization is followed by an infinite loop (steps 2–12) that stops when, for all rules in the beam, it is no longer possible to further improve their quality. Rules can be improved only by conjunctively adding features from $L$. After the first iteration, a rule condition consists of a single feature, after the second iteration up to two features, and so forth. The search is systematic in the sense that for all rules in the beam (step 3) all features from $L$ (step 4) are tested in each iteration. For every new rule, constructed by conjunctively adding a feature to rule body (step 5) quality $q_g$ is computed (step 6). If the support of the new rule is greater than $min\_support$ and if its quality $q_g$ is greater than the quality of any rule in $New\_beam$, the worst rule in $New\_beam$ is replaced by the new rule. The rules are reordered in $New\_beam$ according to their quality $q_g$. At the end of each iteration, $New\_beam$ is copied into $Beam$ (step 11). When the algorithm terminates, the first rule in $Beam$ is the rule with maximum $q_g$.

A necessary condition (in step 7) for a rule to be included in $New\_beam$ is that it must be *relevant*. The new rule is irrelevant if there exists a rule $R$ in $New\_beam$ such that true positives of the new rule are a subset of true positives of $R$ and false positives of the new rule are a superset of false positives of $R$. A detailed analysis of relevance, presented by Lavrač, Gamberger, and Turney (1998), is out of the main scope of this paper. After the new rule is included in $New\_beam$ it may happen that some of the existing rules in $New\_beam$ become irrelevant with respect to this new rule. Such rules are eliminated from $New\_beam$ during its reordering (in step 8). The testing of relevance ensures that $New\_beam$ contains only different and relevant rules.

In Algorithm SD, rule quality measure $q_g$ serves two purposes: first, rule evaluation, and second, evaluation of features and their conjunctions with high potential for the construction of high quality rules in subsequent iterations. The analysis of this quality measure in Section 4 shows that for the first purpose, a measure assigning different costs to false positives and false negatives could perform equally well, but for the purpose of guiding the search the $q_g$ measure is advantageous.

## 2.3 Rule Subset Selection

This section describes how to reduce the number of generated rules to a relatively small number of diverse rules. Reducing the rule set is desirable because expecting experts to evaluate a large set of rules is unfeasible, and second, experiments demonstrate that there are subsets of very similar rules which use almost the same attribute values and have similar prediction properties.

The weighted covering approach proposed for confirmation rule subset selection (Gamberger & Lavrač, 2000) defines diverse rules as those that cover diverse sets of target class examples. The approach, implemented in Algorithm RSS outlined in Figure 2, can not guarantee statistical independence of the selected rules, but it ensures the diversity of generated subsets.

**Algorithm RSS: Rule Subset Selection**
**Input:**    $S$ set of rules for the target class
              $P$ target class examples
**Parameter:** *number* (required number of selected rules in output set $SS$)
**Output:** $SS$ set of relatively independent rules for the target class
(1)    **initialize** $SS \leftarrow \{\}$ (empty set of selected rules)
(2)    **for** every $e \in P$ **do** $c(e) \leftarrow 1$
(3)    **repeat** *number* times
(4)        **select** from $S$ the rule with the highest weight $\sum 1/c(e)$ where summation is
              over the set $P' \subseteq P$ of target class examples covered by the rule
(5)        **for** every $e \in P'$ covered by the selected rule
              **do** $c(e) \leftarrow c(e) + 1$
(6)        **eliminate** the selected rule from $S$
(7)        **add** the selected rule into set $SS$
(8)    **end repeat**

Figure 2: Heuristic rule subset selection algorithm.

Input to Algorithm RSS are the set of all target class examples $P$ and the set of rules $S$. Its output is a reduced set of rules $SS$, $SS \subset S$. The user adjustable parameter *number* determines how many rules will be selected for inclusion in output set $SS$. For every example $e \in P$ there is a counter $c(e)$. Initially, the output set of selected rules is empty (step 1) and all counter values are set to 1 (step 2). Next, in each iteration of the loop (steps 3 to 8), one rule is added to the output set (step 7). From set $S$, the rule with the highest *weight* value is selected. For each rule, *weight* is computed so that $1/c(e)$ values are added for all target class examples covered by this rule (step 4). After rule selection, the rule is eliminated from set $S$ (step 6) and $c(e)$ values for all target class examples covered by the selected rule are incremented by 1 (step 5). This is the central part of the algorithm which ensures that in the first iteration all target class examples contribute the same value $1/c(e) = 1$ to the *weight*, while in the following iterations the contributions of examples are inverse proportional to their coverage by previously selected rules. In this way the examples already covered by one or more selected rules decrease their *weights* while rules covering many yet uncovered target class examples whose weights have not been decreased will have a greater chance to be selected in the following iterations.

In the publicly available Data Mining Server, RSS is implemented in an outer loop for SD. Figure 3 gives the pseudo code of algorithm DMS. In its inner loop, DMS calls SD and selects from its beam the single best rule to be included into the output set $SS$. To enable SD to induce a different solution at each iteration, example weights $c(e)$ are introduced and used in the quality measure which is defined as follows:

$$q_g = \frac{\sum_{TP} \frac{1}{c(e)}}{FP + g}.$$

This is the same quality measure as in SD except that the weights of true positive examples are not constant and equal to 1 but defined by expression $\frac{1}{c(e)}$, changing from iteration to iteration.

The main reason for the described implementation is to ensure the diversity of induced subgroups even though, because of the short execution time limit on the publicly available server, a low *beam_width* parameter value in Algorithm SD had to be set (the default value is 20). Despite the favorable diversity of rules achieved through Algorithm DMS, the approach has also some drawbacks. The first drawback is that the same rule can be detected in different iterations of Algorithm DMS, despite of the changes in the $c(e)$ values. The more important drawback is that heuristic search with a small *beam_width* value may prevent the detection of some good quality subgroups. Therefore during exploratory applications, applying a single SD execution with a large *beam_width* followed by a single run of RSS appears to be a better approach.

## 2.4 Subgroup Search and Evaluation in the ROC and TP/FP Space

The goal of this section is to clarify the relation between the ROC space which is usually used for evaluating classifier performance, and the $TP/FP$ space which is being searched by the $q_g$ heuristic in the SD algorithm.

Evaluation of induced subgroups in the ROC space (ROC: Receiver Operating Characteristic, Provost & Fawcett, 2001) shows their performance in terms of $TPr$ and $FPr$,

**Algorithm DMS: Data Mining Server subgroup construction**
**Input:**   $E = P \cup N$ ($E$ training set, $|E|$ training set size,
                    $P$ positive (target class) examples,
                    $N$ negative (non-target class) examples)
         $L$ set of all defined features (attribute values), $l \in L$
**Parameter:** *number* (required number of selected rules
             in output set $SS$)
             $g$ (generalization parameter, $0.1 < g < 100$, default value 1)
             *min_support* (minimal support for rule acceptance)
             *beam_width* (number of rules in the beam)
**Output:** $SS$ set of relatively independent rules for the target class
(1)     **initialize** $SS \leftarrow \{\}$ (empty set of selected rules)
(2)     **for** every $e \in P$ **do** $c(e) \leftarrow 1$
(3)     **repeat** *number* times
(4)         call **Algorithm SD** to construct a rule with maximal
             quality $q_g = \frac{\sum_{TP} \frac{1}{c(e)}}{FP + g}$
(5)         **for** every $e \in P'$ covered by the constructed rule
                 **do** $c(e) \leftarrow c(e) + 1$
(6)         **add** the constructed rule into set $SS$
(7)     **end repeat**

Figure 3: Iterative subgroup construction in the Data Mining Server.

where $TPr$ is the *sensitivity* of a classifier measuring the fraction of positive cases that are classified as positive, and $FPr$ is the *false alarm* measuring the fraction of incorrectly classified negative cases: $TPr = \frac{TP}{TP+FN} = \frac{TP}{Pos}$, and $FPr = \frac{FP}{TN+FP} = \frac{FP}{Neg}$. A point in the ROC space shows classifier performance in terms of false alarm rate $FPr$ (plotted on the $X$-axis) that should be as low as possible, and sensitivity $TPr$ (plotted on the $Y$-axis) that should be as high as possible (see Figure 5 in Section 3.2).

The ROC space is appropriate for measuring the success of subgroup discovery, since subgroups whose $TPr/FPr$ tradeoff is close to the diagonal can be discarded as uninteresting. Conversely, interesting rules/subgroups are those sufficiently distant from the diagonal. Those rules which are most distant from the diagonal define the points in the ROC space from which a convex hull is constructed. The area under the ROC curve defined by subgroups with the best $TPr/FPr$ tradeoff can be used as a quality measure for comparing the success of different learners or subgroup miners. In subgroup construction, the data analyst can try to achieve the desired $TPr/FPr$ tradeoff by building rules using different data mining algorithms, by different parameter settings of a selected data mining algorithm or by applying a cost-sensitive data mining algorithm that takes into the account different misclassification costs.

The $q_g$ measure in the SD algorithm that needs to be maximized, tries to find subgroups that are as far as possible from the diagonal of the ROC space in the direcion of the left upper corner (with $TPr$ equal to 100% and $FPr$ equal to 0%). Note, however, that the actual computation, as implemented in Algorithm SD, is not performed in terms of $TPr$ and $FPr$, as assumed in the ROC analysis, but rather in terms of $TP$ and $FP$ in the so-called $TP/FP$ space. The reason is the improved computational efficiency of computing the $q_g$ value which is used as a search heuristic for comparing the quality of rules for a given, fixed domain. For a fixed domain, the $TP/FP$ space is as appropriate as the ROC space: the ROC space is namely equivalent to the normalized $TP/FP$ space where $Pos$ and $Neg$ are normalization constants for $Y$ and $X$ axes, respectively. The $TP/FP$ space and the ROC space are illustrated in Section 3.2 by Figures 4 and 5, respectively.

## 3. The Descriptive Induction Process

The induction of subgroups, described in Section 2.2, represents the main step of the proposed descriptive induction process. This step corresponds to the data mining step of the standard process of knowledge discovery in databases (KDD). The overall descriptive induction process, proposed in this paper, is comparable to the standard KDD process (Fayyad, Piatetsky-Shapiro, & Smyth, 1996), with some particularities of the task of subgroup discovery.

The proposed expert-guided subgroup discovery process consists of the following steps:

1. problem understanding

2. data understanding and preparation

3. subgroup detection

4. subgroup subset selection

5. statistical characterization of subgroups

6. subgroup visualization

7. subgroup interpretation

8. subgroup evaluation

Section 3.1, illustrating steps 1 and 2, presents a medical problem used as a case study for applying the proposed descriptive induction methodology. Tools for supporting subgroup detection and selection in steps 3 and 4 were described in detail in Sections 2.2 and 2.3, while the results of expert-guided subgroup detection and selection are outlined in Section 3.2. Methods and results of steps 5–8 for this domain are outlined in Sections 3.3–3.6, respectively.

The proposed descriptive induction process is iterative and interactive. It is iterative, since many steps may need to be repeated before a satisfactory solution is found. It is also interactive, assuming expert's involvement in most of the phases of the proposed descriptive induction process. The expert's role in the patient risk group detection application is described in Section 3.7.

## 3.1 The Problem of Patient Risk Group Detection

Early detection of artherosclerotic coronary heart disease (CHD) is an important and difficult medical problem. CHD risk factors include artherosclerotic attributes, living habits, hemostatic factors, blood pressure, and metabolic factors (Goldman et al., 1996). Their screening is performed in general practice by data collection in three different stages.

**A** Collecting anamnestic information and physical examination results, including risk factors like age, positive family history, weight, height, cigarette smoking, alcohol consumption, blood pressure, and previous heart and vascular diseases.

**B** Collecting results of laboratory tests, including information about risk factors like lipid profile, glucose tolerance, and trombogenic factors.

**C** Collecting ECG at rest test results, including measurements of heart rate, left ventricular hypertrophy, ST segment depression, cardiac arrhythmias and conduction disturbances.

Our goal was to construct at least one relevant and interesting subgroup, called a pattern in the rest of the work, for each stage, A, B, and C, respectively.

A database with 238 patients representing typical medical practice in CHD diagnosis, collected at the Institute for Cardiovascular Prevention and Rehabilitation, Zagreb, Croatia, was used for subgroup discovery. The database is in no respect a good epidemiological CHD database reflecting actual CHD occurrence in a general population, since about 50% of gathered patient records represent CHD patients. Nevertheless, the database is very valuable since it includes records of different types of the disease. Moreover, the included negative cases (patients who do not have CHD) are not randomly selected persons but individuals with some subjective problems or those considered by general practitioners as potential CHD patients, and hence sent for further investigations to the Institute. This biased data set is appropriate for CHD risk group discovery, but it is inappropriate for measuring the success of CHD risk detection and for subgroup performance estimation in general medical practice.

## 3.2 Results of Expert-Guided Subgroup Detection and Selection

The process of expert-guided subgroup discovery was performed as follows. For every data stage A, B and C, the DMS algorithm was run for values $g$ in the range 0.5 to 100, and a fixed number of selected output rules equal to 3. The rules induced in this iterative process were shown to the expert for selection and interpretation. The inspection of 15–20 rules for each data stage triggered further experiments. Concrete suggestions of the medical expert involved in this study were to limit the number of features in the rule body and to try to avoid the generation of rules whose features would involve expensive and/or unreliable laboratory tests. Consequently, we have performed the further experiments by intentionally limiting the feature space and the number of iterations in the main loop of the SD algorithm (steps 2-12 of Algorithm SD).

In this iterative process, the expert has selected five interesting CHD risk groups. Table 1 shows the induced subgroups, together with the values of $g$ and the rule significance. In the subgroup discovery terminology proposed in this paper, the features appearing in the conditions of rules describing the subgroups are called the *principal factors*. The described iterative process was successful for data at stages B and C, but it turned out that anamnestic data on its own (stage A data) is not informative enough for inducing subgroups, i.e., it failed to fulfil the expert's criteria of interestingness. Only after engineering the domain, by separating male and female patients, were interesting subgroups discovered. See Section 3.7 for more details on the expert's involvement in this subgroup discovery process.

| | | | **Expert Selected Subgroups** | $g$ | *Sig* |
|---|---|---|---|---|---|
| A1 | CHD | ← | positive family history **AND** age over 46 year | 14 | 95% |
| A2 | CHD | ← | body mass index over 25 $kgm^{-2}$ **AND** age over 63 years | 8 | 99% |
| B1 | CHD | ← | total cholesterol over 6.1 $mmolL^{-1}$ **AND** age over 53 years **AND** body mass index below 30 $kgm^{-2}$ | 10 | 99.9% |
| B2 | CHD | ← | total cholesterol over 5.6 $mmolL^{-1}$ **AND** fibrinogen over 3.7 $gL^{-1}$ **AND** body mass index below 30 $kgm^{-2}$ | 12 | 99.9% |
| C1 | CHD | ← | left ventricular hypertrophy | 10 | 99.9% |

Table 1: Induced subgroups in the form of rules. Rule conditions are conjunctions of principal factors. Subgroup A1 is for male patients, subgroup A2 for female patients, while subgroups B1, B2, and C1 are for male and female patients. The subgroups are induced from different attribute subsets with corresponding $g$ parameter values given in column $g$. The last column *Sig* contains information about the significance of the rules computed by the $\chi^2$ test.

Separately for each data stage, we have investigated which of the induced rules are the best in terms of the ROC space, i.e., which of them are used to define the ROC convex hull. At stage B, for instance, seven rules are on the convex hull shown in Figures 4 and 5 for the $TP/FP$ and the ROC space, respectively. Two of these rules, $X1$ and $X2$, indicated in the figures, are listed in Table 2. Notice that the expert-selected subgroups B1 and B2 are significant, but are not among those lying on the convex hull. The reason for selecting exactly those two rules at stage B are their simplicity (consisting of three features only), their generality (covering relatively many positive cases) and the fact that the used features are, from the medical point of view, inexpensive laboratory tests.

### 3.3 Statistical Characterization of Subgroups

The next step in the proposed descriptive induction process starts from the discovered subgroups. In this step, statistical differences in distributions are computed for two populations, the target and the reference population. The target population consists of true positive case (CHD patients included into the analyzed subgroup), whereas the reference population are all available non-target class examples (all the healthy subjects).

| | | | Best Induced Subgroups | $g$ | *Sig* |
|---|---|---|---|---|---|
| X1 | CHD | ← | age over 61 years **AND** tryglicerides below 1.85 $mmolL^{-1}$ **AND** high density lipoprotein below 1.25 $mmolL^{-1}$ | 4 | 99.9% |
| X2 | CHD | ← | body mass index over 25 **AND** high density lipoprotein below 1.25 $mmolL^{-1}$ **AND** uric acid below 360 $mmolL^{-1}$ **AND** glucose below 7 $mmolL^{-1}$ **AND** fibrinogen over 3.7 $gL^{-1}$ | 16 | 99.9% |

Table 2: Two of the best induced subgroups induced for stage B. Their position in the $TP/FP$ and the ROC space are marked in Figures 4 and 5, respectively.

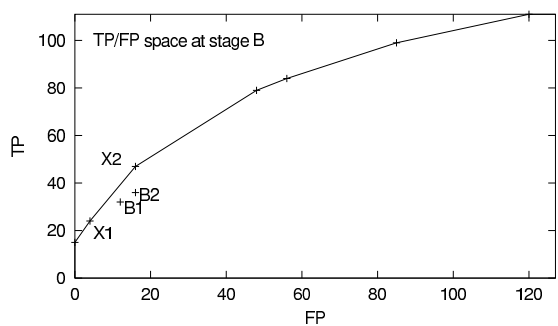

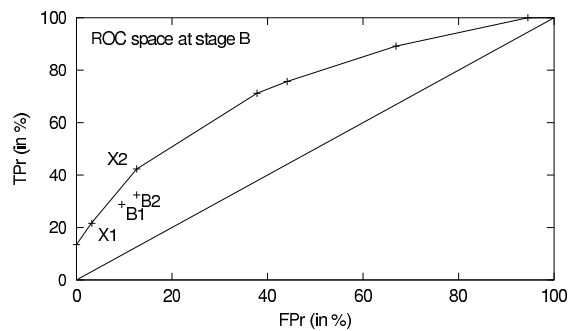

Figure 4: The $TP/FP$ space presenting the convex hull of subgroups induced using the quality measure $q_g = TP/(FP + g)$ at data stage B. Labels B1 and B2 denote positions of subgroups selected by the medical expert, and X1 and X2 two of the seven subgroups forming the $TP/FP$ convex hull.

Figure 5: The same subgroups as in Figure 4 shown in the ROC space instead of the $TP/FP$ space. The equivalence of these two spaces can be easily noticed. In the ROC space a thin line connecting points (0,0) and (100,100) represents rule positions with significance equal zero.

Statistical differences in distributions for all the descriptors (attributes) between these two populations is tested using the $\chi^2$ test with 95% confidence stage ($p = 0.05$). For this purpose numerical attributes have been partitioned in up to 30 intervals so that in every interval there are at least 5 instances. Among the attributes with significantly different distributions there are always those that form the features describing the subgroups (the principal factors), but usually there are also other attributes with significantly different value distributions. These attributes are called *supporting attributes*, and the features formed of their values that are characteristic for the discovered subgroups are called *supporting factors*.

Supporting factors are very important to achieve pattern descriptions that are reasonably complete and acceptable for medical practice, as medical experts dislike short rules and prefer rules including as much supportive evidence as possible (Kononenko, 1993).

In this work, the role of statistical analysis is to detect meaningful supporting factors, whereas the decision whether they will be used to support user's confidence in the subgroup description is left to the expert. In the CHD application the expert has decided whether the proposed factors are indeed interesting, how reliable they are or how easily they can be measured in practice. In Table 3, expert selected supporting factors are listed next to the individual CHD risk groups, each described by a list of principal factors.

| | Principal Factors | Supporting Factors |
|---|---|---|
| A1 | positive family history<br>age over 46 year | psychosocial stress<br>cigarette smoking<br>hypertension<br>overweight |
| A2 | body mass index over 25 $kgm^{-2}$<br>age over 63 years | positive family history<br>hypertension<br>slightly increased LDL cholesterol<br>normal but decreased HDL cholesterol |
| B1 | total cholesterol over 6.1 $mmolL^{-1}$<br>age over 53 years<br>body mass index below 30 $kgm^{-2}$ | increased triglycerides value |
| B2 | total cholesterol over 5.6 $mmolL^{-1}$<br>fibrinogen over 3.7 $mmolL^{-1}$<br>body mass index below 30 $kgm^{-2}$ | positive family history |
| C1 | left ventricular hypertrophy | positive family history<br>hypertension<br>diabetes mellitus |

Table 3: Induced subgroup descriptions (principal factors) and their statistical characterizations (supporting factors).

## 3.4 Subgroup Visualization

A novel visualization method can be used to visualize the output of any subgroup discovery algorithm, provided that the output has the form of rules with a target class in their consequent. It can also be used as a method for visualizing standard classification rules.

Subgroup visualization, as described in this section, allows us to compare distributions of different subgroups. The approach assumes the existence of at least one numeric (or ordered discrete) attribute of expert's interest for subgroup analysis. The selected attribute is plotted on the $X$-axis of the diagram. The $Y$-axis represents a class, or more precisely, the number of instances of a given class. Both directions of the $Y$-axis ($Y^+$ and $Y^-$) are used to indicate the number of instances. In Figure 6, for instance, the $X$-axis represents *age*, the $Y^+$-axis denotes class coronary heart disease (CHD) and $Y^-$ denotes class 'healthy' (non-CHD). Out of four graphs at the $Y^+$ side, three represent induced subgroups ($A1$, $A2$ and $C1$) of CHD patients, and the fourth shows the age distribution of the entire population of CHD (all CHD) patients. The graphs at the $Y^-$ side show the distribution of non-CHD (all healthy) patients in the training set and the distribution of healthy subjects included into the subgroup A2 (dashed line).

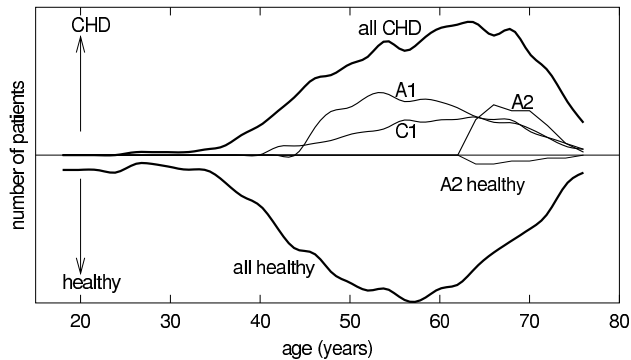

Figure 6: Distributions of the numbers of CHD patients (all CHD) and healthy subjects (all healthy) in terms of age (in years). Graphs $A1$, $A2$, and $C1$ represent the distributions of CHD patients belonging to the corresponding subgroups. The dashed line represents healthy subjects included in subgroup $A2$.

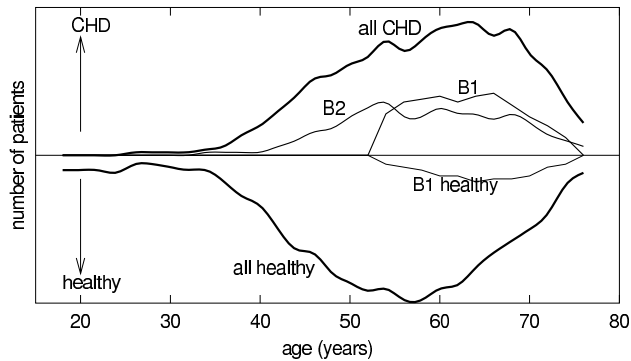

Figure 7: Distributions of the numbers of CHD patients (all CHD) and healthy subjects (all healthy), as well as the distributions of patients for subgroups $B1$ and $B2$ in terms of age (in years). The dashed line represents healthy subjects included in subgroup $B1$.

On purpose, the graphs of subgroups $A1$ and $C1$ in Figure 6 show only the coverage of positive cases (CHD patients), and in Figure 7 the graph of subgroup $B2$ shows only the coverage of positive cases, whereas the graphs of $A2$ in Figure 6 and $B1$ in Figure 7 indicate that the descriptions of subgroups cover positive cases (CHD patients) as well as some negative cases (healthy individuals). Except for the correct visualization of subgroups $A2$ and $B1$ and of the entire CHD and non-CHD distribution, Figures 6 and 7 have been simplified in order to enable a better understanding of the visualization method, by showing just the coverage of positive cases.

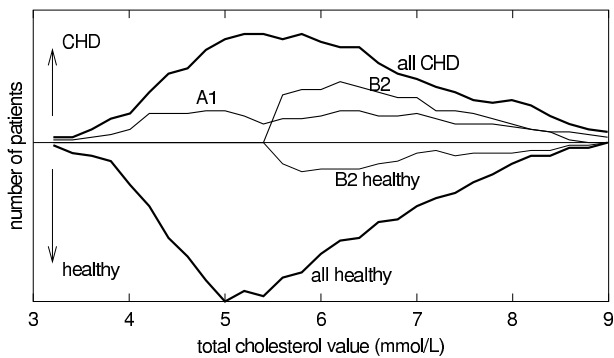

Figure 8: Distributions of all CHD patients and those described by patterns A1 and B2, as well as all healthy subjects and those included into pattern B2 (dashed line) in terms of total cholesterol value in $mmol\,L^{-1}$.

In medical domains we typically use the $Y^+$ side to represent the number of positive cases (CHD patients, in this paper) in order to reveal properties of induced patterns for subgroups of these patients. On the other hand, the $Y^-$ side is reserved to reveal properties of these same patterns (or other patterns) for the negative cases (patients without CHD). One of the advantages of using $Y^+$ and $Y^-$ as proposed above is that in binary classification problems the comparison of the area under the graph of a subgroup and the graph of the entire population visualizes the fractions of $\frac{TP}{Pos} = \frac{TP}{TP+FN}$ at the $Y^+$ side (sensitivity $TPr$), and $\frac{FP}{Neg} = \frac{FP}{TN+FP}$ at the $Y^-$ side (false alarm rate $FPr$), where $Pos$ and $Neg$ stand for the numbers of positive and negative cases in the entire population, respectively. For instance, in the visualization of subgroup $B1$ in Figure 7 the area under the dashed line on the $Y^-$ side represents the numbers of misclassified training instances of subgroup $B1$. In this way, the sensitivity and false alarm rate can be estimated for pattern $B1$ from Figure 7. The same information for pattern B2 can be found in Figure 8, showing subgroups A1 and B2 in terms of attribute 'total cholesterol value'.

The proposed visualization method can be adapted to visualize subgroups also in terms of value distributions of discrete/nominal attributes. An approach to such visualization is presented in Figure 9. However, due to bar chart representation, it is more difficult to compare several subgroups in one visualization.

In general, it is not necessary that $Y^+$ and $Y^-$ denote two opposite classes. If appropriate, they may denote any two classes, or even any two different attribute values, which the expert would like to compare.

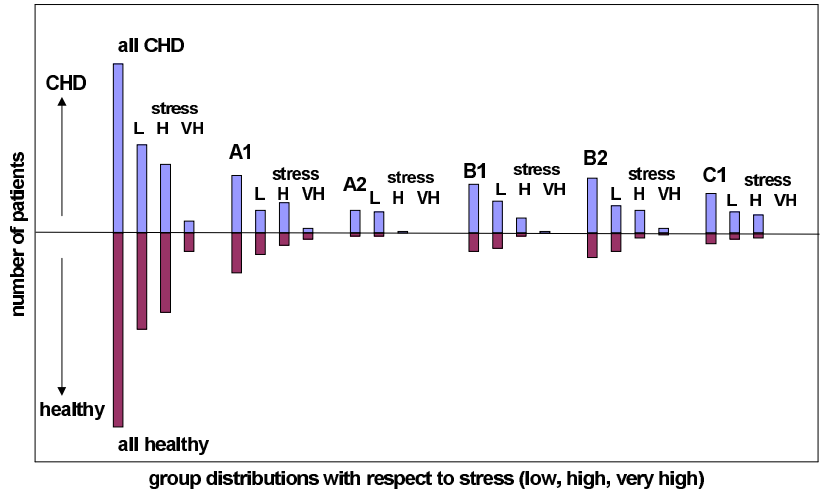

Figure 9: Distribution of CHD patients and healthy subjects with respect to stress values (low, high, and very high) for the entire population and the five induced patterns.

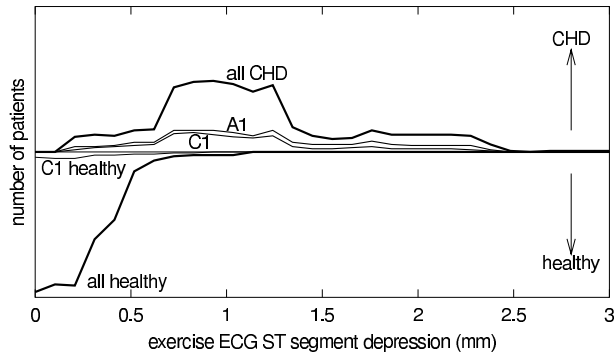

Figure 10: Distribution of CHD patients and healthy subjects with respect to exercise ECG ST segment depression in millimeters (1mm corresponds to 0.1 mV). Large difference between total healthy and ill populations can be noticed, but differences among patterns are very small. Patterns A1 and C1 are selected as extreme cases. The dashed line presents healthy persons incorrectly described by pattern C1

## 3.5 Subgroup Interpretation through Visualization

Subgroup visualization is very valuable for expert interpretation of subgroup discovery results. From Figures 6 and 7 it can be seen that there is no significant difference between CHD patients and healthy subjects regarding their age, but that there are significant differences among the detected patterns. Figure 8 illustrates a similar effect for the total cholesterol values although it is known that total cholesterol is an important risk factor for the CHD disease. This observation shows that the problem of CHD risk group detection can typically not be solved by considering single features and demonstrates the appropriateness of the suggested approach which tries to generate subgroup descriptions which are a logical conjunction of a few correlated features.

Figure 10 is also interesting, since it is very different from other figures. Notice that exercise ECG ST segment depression was not used as an attribute in the training data (which contained only attributes that are available at stages A, B and C); exercise ECG ST segment depression, long term ECG recording and echochardiography are not available for early risk group detection since they can be collected/measured only in specialized medical institutions. Figure 10 clearly demonstrates significant differences between all CHD and all healthy subjects in terms of exercise ECG ST segment depression values, demonstrating that this measurement, if available, is an excellent disease indicator. But it also shows that, although it is known that patterns A1 and C1 cover different disease subpopulations, they behave very similarly in terms of the exercise ECG ST segment depression property.

## 3.6 Subgroup Evaluation

In order to evaluate the discovered risk groups, the medical expert has tested the induced subgroup patterns on an independent set of 70 people (50 CHD patients and 20 non-CHD cases from the same hospital). The results for these patients, summarized in Table 4, show that the patterns are successful in detecting CHD patients. About 90% of CHD patients were included into at least one of the five patterns. The detected sensitivity values ($TPr$) for patterns A1, B2, and C1 are significantly higher than the values computed on the set of patients used for subgroup discovery. For the other two patterns the values do not differ significantly. Note that the accuracy values are relatively high, despite the relatively high false positive rate ($FPr$): a lower $FPr$ could have been achieved by selecting lower values of the generalization parameter $g$, at a cost of detecting subgroups with lower coverage of positive cases.

|    | Training set | | | Test set | | |
|----|-------|-------|----------|-------|-------|----------|
|    | $TPr$ | $FPr$ | $Accuracy$ | $TPr$ | $FPr$ | $Accuracy$ |
| A1 | 47.5% | 26.8% | 59.4% | 84.8% | 77.8% | 80.0% |
| A2 | 48.4% | 6.7%  | 81.2% | 41.2% | 27.3% | 70.0% |
| B1 | 28.8% | 9.4%  | 72.7% | 36.0% | 20.0% | 81.8% |
| B2 | 32.4% | 12.6% | 69.2% | 42.0% | 15.0% | 87.5% |
| C1 | 23.4% | 5.5%  | 78.8% | 82.0% | 40.0% | 83.7% |

Table 4: Summary of results obtained on the training set and on an independent set of 70 persons (50 CHD patients and 20 non-CHD cases from the same hospital), measured in terms of $TPr$, $FPr$ and $Accuracy$.

### 3.7 The Expert's Role in Subgroup Discovery

The CHD case study illustrates that expert-guided induction is an iterative process in which the expert can change the requested generality of the induced subgroups and the subset of attributes (features) that are made available for rule construction. In this way it is possible to induce different patterns (subgroups) from the same data set. The selection of one or more subgroups representing the final solution is left to the expert; the decision depends both on rule prediction properties (like the number of true positives and the tolerated number of false positives), as well as subjective properties like the understandability, unexpectedness and actionability of induced subgroup descriptions (Silberschatz & Tuzhilin, 1995), which depend on the features used in the conditions of induced rules. In the application described in this paper, the main subjective acceptability criteria were understandability, simplicity and actionability.

Partitioning the CHD risk group problem into three data stages A–C was completely based on the expert's understanding of the typical diagnostic process. From the machine learning point of view this affects the selection of subsets of attributes that are used in different experiments. Moreover, at data stage A the partitioning of the example set has been used as well. At this data stage there are only a few attributes that could have been used for rule induction. The expert's understanding of the domain suggested that the CHD population be partitioned into two subpopulations based on the sex of patients, making it significantly easier to induce interesting subgroups. This partitioning resulted in patterns A1 and A2.

Alternatively, partitioning can be performed also in the phase of performing statistical characterization of discovered subgroups, by further splitting the detected subgroups in several parts (e.g., differentiating between male and female patients that are true positive cases for the subgroup) and then comparing attribute value distributions for these parts. Any significant difference in this distribution may be potentially interesting as part of the subgroup description. As a basis for subgroup partitioning one may use either some detected supporting risk factor or any other attribute or attribute combination which is potentially interesting based on the existing expert knowledge.

There has been some effort devoted also to automating the process of partitioning example sets by a method of unsupervised learning, but its presentation is out of the main scope of this work (Šmuc, Gamberger, & Krstačić, 2001).

From the methodological point of view it is interesting to notice that the expert appreciated the induced subgroups covering many target class cases (with true positive rate of at least 20%) and with false positive rate as low as possible, with the intention to keep it below 10%. But in selecting a rule, its prediction quality has not been the most important factor. The necessary condition for selecting a rule was that the expert was able to recognize connections among features building the rule that are medically reasonable. In this sense, short rules are significantly more intuitive; it can be noticed from Table 1 that all rules selected by the expert have at most three features defining the principal risk factors. The fact that the expert did not select subgroups with an optimal $TP/FP$ ratio is illustrated by Figures 16–18 in Section 4.2, which show the positions of the patterns A1–C1 in the $TP/FP$ space and the $TP/FP$ convex hulls induced for data stages A–C, connecting points with the optimal coverage properties. It can be noticed that none of the expert selected patterns is lying on the $TP/FP$ convex hull but the selected patterns are close to the convex hull.

## 4. Analysis of the Proposed Rule Quality Measure Used in Heuristic Search

It is well known from the ROC analysis, that in order to achieve the best results, the discovered rules should be as close as possible to the top-left corner of the ROC space. This means that in the $TPr/FPr$ tradeoff, $TPr$ should be as large as possible, and $FPr$ as small as possible. Similarly, in the $TP/FP$ space, $TP$ should be as large as possible, and $FP$ as small as possible.

In this work the quality measure $q_g = TP/(FP + g)$ using generalization parameter $g$ has been defined. This section explains why this quality measure has been selected, in comparison with other more intuitive quality measure like a cost-based measure $q_c$ involving 'cost' parameter $c$.

### 4.1 Comparison of the $q_g$ and $q_c$ Heuristics

Our experience in different medical applications indicates that intuitions like "how expensive is every $FP$ prediction in terms of additional $TP$ predictions made by a rule" are useful for understanding the problem of directing the search in the $TP/FP$ space. Suppose that the definition of cost parameter $c$ is based on the following argument: "For every additional $FP$, the rule should cover more than $c$ additional $TP$ examples in order to be better." Based on such reasoning, it is possible to define a quality measure $q_c$, using the following $TP/FP$ tradeoff:

$$q_c = TP - c \cdot FP.$$

Quality measure $q_c$ is easy to use because of the intuitive interpretation of parameter $c$. It has also a nice property when used for subgroup discovery: by changing the $c$ value we can move in the $TP/FP$ space and select the optimal point based on parameter $c$.

In Algorithm SD, the quality measure $q_g$, using a different $TP/FP$ tradeoff is used: $q_g = TP/(FP + g)$, where $g$ is the generalization parameter.

If a subgroup discovery algorithm employs exhaustive search (or if all points in the $TP/FP$ space are known in advance) then the two measures $q_g$ and $q_c$ are equivalent in the sense that every optimal solution lying on the convex hull can be detected by using any of the two heuristics; only the values that must be selected for parameters $g$ and $c$ are different. In this case, $q_c$ might be even better because its interpretation is more intuitive.

However, since Algorithm SD is a heuristic beam search algorithm, the situation is different. Subgroup discovery is an iterative process, performing one or more iterations (typically 2–5) until good rules are constructed by forming conjunctions of features in the rule body. In this process, a rule quality measure is used for rule selection (for which the two measures $q_g$ and $q_c$ are equivalent) as well as for the selection of features and their conjunctions that have high potential for the construction of high quality rules in subsequent iterations; for this use, rule quality measure $q_g$ is better than $q_c$. Let us explain why.

Suppose that we have a point (a rule) $R$ in the $TP/FP$ space, where $TP$ and $FP$ are its true and false positives, respectively. For a selected $g$ value, $q_g$ can be determined for this rule $R$. It can be shown that all points that have the same quality $q_g$ as rule $R$ lie on a line defined by the following function:

$$tp = \frac{TP \cdot fp}{FP + g} + \frac{TP \cdot g}{FP + g} = \frac{TP \cdot (fp + g)}{FP + g}.$$

In this function, $tp$ represents the number of true positives of a rule with quality $q_g$ which covers exactly $fp$ negative examples. By selecting a different $fp$ value, the corresponding $tp$ value can be determined by this function. The line, determined by this function, crosses the $tp$ axis at point $TP_0 = TP \cdot g/(FP + g)$ and the $fp$ axis at point $-g$. This is shown in Figure 11. The slope of this line is equal to the quality of rule $R$, which equals $TP/(FP+g)$.

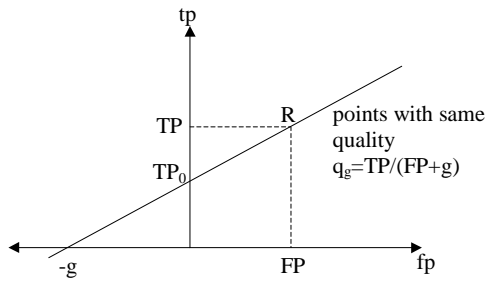

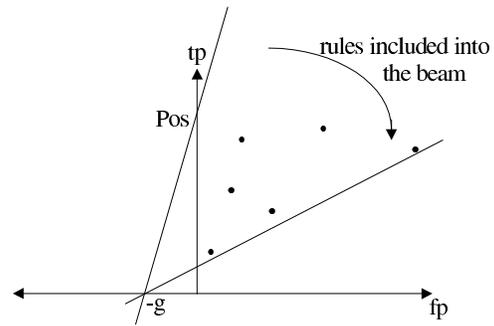

Figure 11: Properties of rules with the same quality $q_g$.

Figure 12: Rules with highest quality included into the beam for $q_g = TP/(FP + g)$.

In the $TP/FP$ space, points with higher quality than $q_g$ are above this line, in the direction of the upper left corner. Notice that in the $TP/FP$ space the top-left is the preferred part of the space: points in that part represent rules with the best $TP/FP$ tradeoff. This reasoning indicates that points that will be included in the beam must all lie above the line of equal weights $q_{beam}$ which is defined by the last rule in the beam. If represented graphically, first $beam\_width$ number of rules, found in the $TP/FP$ space when rotating the line from point $(0, Pos)$ in the clockwise direction, will be included in the beam. The center of rotation is point $(-g, 0)$. This is illustrated in Figure 12.

On the other hand, for the $q_c$ quality measure defined by $q_c = TP - c \cdot FP$ the situation is similar but not identical. Points with the same quality lie on a line $tp = c \cdot (fp - FP) + TP$, but its slope is constant and equal to $c$. Points with higher quality lie above the line in the direction of the left upper corner. The points that will be included into the beam are the first $beam\_width$ points in the $TP/FP$ space found by a *parallel* movement of the line with slope $c$, starting from point $(0, Pos)$ in the direction towards the lower right corner. This is illustrated in Figure 13.

Let us now assume that we are looking for an optimal rule which is very specific. In this case, parameter $c$ will have a high value while parameter $g$ will have a very small value. The intention is to find the same optimal rule in the $TP/FP$ space. At the first stage of rule construction only single features are considered and most probably their quality as the final solution is rather poor. See Figure 14 for a typical placement of potentially interesting features in the $TP/FP$ space.

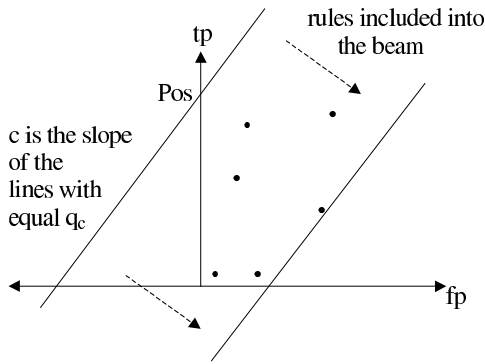

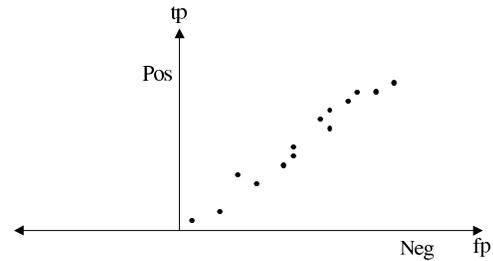

Figure 13: Rules with highest quality included in the beam for $q_c = TP - c \cdot FP$.

Figure 14: Placement of interesting features in the $TP/FP$ space after the first iteration.

The primary function of these features is to be good building blocks so that by conjunctively adding other features, high quality rules can be constructed. By adding conjunctions, solutions generally move in the direction of the left lower corner. The reason is that conjunctions can reduce the number of $FP$ predictions. However, they reduce the number of $TP$'s as well. Consequently, by conjunctively adding features to rules that are close to the left lower corner, the algorithm will not be able to find their specializations nearer to the left upper corner. Only the rules that have high $TP$ value, and are in the upper part of the $TP/FP$ space, have a chance to take part in the construction of interesting new rules.

Figure 15 illustrates the main difference between quality measures $q_g$ and $q_c$: the former tends to select more general features from the right upper part of the $TP/FP$ space (points in the so-called '$g$ space'), while the later 'prefers' specific features from the left lower corner (points in the so-called '$c$ space'). In cases when $c$ is very large and $g$ is very small, the effect can be so important that it may prevent the algorithm from finding the optimal solution even with a large beam width. Notice, however, that Algorithm SD is heuristic in its nature and no statements are true for all cases. This means that in some, but very rare cases, a quality measure based on parameter $c$ may result in a better final solution.

## 4.2 Experimental Evaluation of the Heuristics

For the purpose of comparing the $q_g$ and $q_c$ measures, a $TP/FP$ convex hull for each of the two measures has been constructed. The procedure was repeated for stages A–C. The $TP/FP$ convex hulls for the $q_g$ measure were constructed so that for different $g$ values many subgroups were constructed. Among them those lying on the convex hull in the $TP/FP$ space were selected: this resulted in convex hulls presented by the thick lines in Figures 16–18. The thin lines represent the $TP/FP$ convex hulls obtained in the same way for subgroups induced by the $q_c$ measure, for $c$ values between 0.1 and 50.

Figures 16-18 for stages A–C demonstrate that both curves agree in the largest part of the $TP/FP$ space, but that for small $FP$ values the $q_g$ measure is able to find subgroups

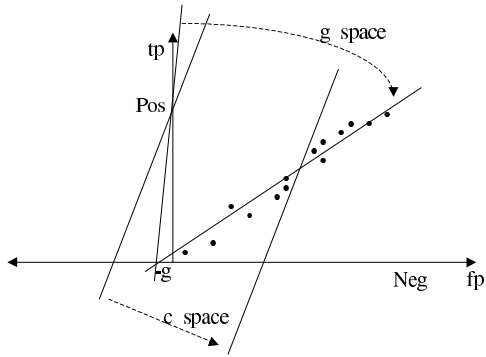

Figure 15: The quality $q_c$ employing the $c$ parameter tends to select patterns (points) with small $TP$ values, while quality $q_g$ employing the $g$ parameter will include also many patterns with large $TP$ values (from the right part of the $TP/FP$ space) that have a chance to be used in building conjunctions of high quality rules.

covering more positive examples. According to the analysis in the previous section, this was the expected result. In order to make the difference more obvious only the left part of the $TP/FP$ space is shown in these figures.

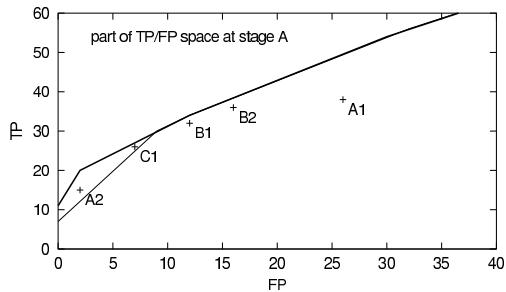

Figure 16: The left part of the $TP/FP$ space presenting the $TP/FP$ convex hulls of subgroups induced using quality measures $q_g = TP/(FP + g)$ (thick line) and $q_c = TP - c \cdot FP$ (thin line) at data stage A. Labels A1–C1 denote positions of subgroups selected by the medical expert as interesting risk group descriptions.

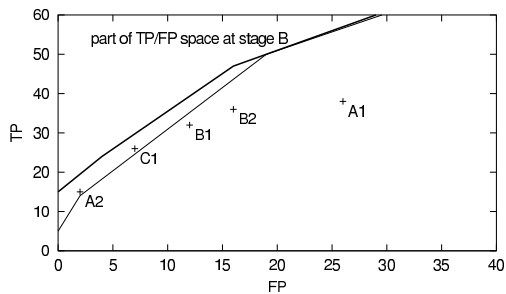

Figure 17: The left part of the $TP/FP$ convex hulls representing subgroups induced at data stage B.

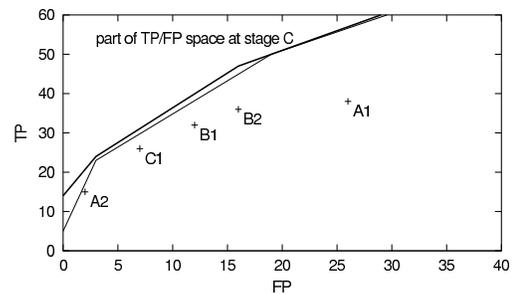

Figure 18: The left part of the $TP/FP$ convex hulls representing subgroups induced at data stage C.

The differences between the $TP/FP$ convex hulls for $q_g$ and $q_c$ measures may seem small and insignificant, but in reality it is not so. The majority of interesting subgroups (this claim is supported also by patterns A1–C1 selected by the domain expert) are subgroups with a small false positive rate which lie in the range in which $q_g$ works better. In addition, for subgroups with $FP = 0$ the true positive rate in our examples was about two times larger for subgroups induced with $q_g$ than with $q_c$. Furthermore, note that for stages A and B there are two out of five subgroups (A2 and C1) which lie in the gap between the $TP/FP$ convex hulls. If the $q_c$ measure instead of $q_g$ measure were used in the experiments with CHD domain, at least subgroup A2 could not have been detected.

## 5. Related Work

This section provides comparisons and links to related work in subgroup discovery, measures of interestingness, evaluation measures and visualization.

### 5.1 Subgroup Discovery

The need for user interactivity in subgroup discovery is addressed by Wrobel S. et al. (1996), describing a system developed in the KESO European research project (Knowledge Extraction for Statistical Offices) and in the systems EXPLORA (Klösgen, 1996) and MIDOS (Wrobel, 1997, 2001). EXPLORA treats the learning task as a single relation problem, i.e., all the data are assumed to be available in one table (relation), whereas MIDOS extends this task to multi-relation databases, which is related to a number of other learning tasks (De Raedt & Dehaspe, 1997; Mannila & Toivonen, 1996; Wrobel & Džeroski, 1995), mostly in the field of Inductive Logic Programming (Džeroski & Lavrač, 2001; Lavrač & Džeroski, 1994).

The most important features of EXPLORA and MIDOS, related to this paper, concern the use of heuristics for subgroup discovery; the measures of interestingness and the search heuristics are outlined in separate sections below. A related approach to our approach to rule subset selection, presented in Section 2.3, is Gebhardt's (1991) work on subgroup suppression.

Note that some approaches to association rule induction can also be used for subgroup discovery. For instance, the APRIORI-C algorithm (Jovanoski & Lavrač, 2001), which applies association rule induction to classification rule induction, outputs classification rules with guaranteed support and confidence with respect to a target class. If a rule satisfies also a user-defined significance threshold, an induced APRIORI-C rule is an independent 'chunk' of knowledge about the target class, which can be viewed as a subgroup description with guaranteed significance, support and confidence. Similarly, the confirmation rule concept, introduced by Gamberger and Lavrač (2000) and used as a basis for the subgroup discovery algorithm in this paper, utilizes the minimal support requirement as a measure which must be satisfied by every rule in order to be included in the induced confirmation rule set.

Both above mentioned approaches to subgroup discovery exploit the information about class membership. One of the main reasons why these approaches are of interest for subgroup discovery is that, unlike the classical classification rule induction algorithms such as CN2 (Clark & Niblett, 1989) and AQ (Michalski, Mozetič, Hong, & Lavrač, 1986), they do not use the covering algorithm. In covering algorithms only the first few induced rules may be of interest as subgroup descriptors with sufficient coverage. Subsequently induced rules are induced from biased example subsets, e.g., subsets including only positive examples not covered by previously induced rules. This bias constrains the population for subgroup discovery in a way that is unnatural for the subgroup discovery process which is, in general, aimed at discovering interesting properties of subgroups of the entire population.

Recent approaches to subgroup discovery aim at overcoming the problem of this inappropriate bias of the standard covering algorithm. The recently developed subgroup discovery algorithms CN2-SD (Lavrač, Flach, Kavšek, & Todorovski, 2002) and RSD (Lavrač, Železný, & Flach, 2002) use the so-called weighted covering algorithm, similar to the one implemented in Algorithm DMS described in this paper.

Instance weights play an important role in boosting (Freund & Shapire, 1996) and alternating decision trees (Schapire & Singer, 1999). Instance weights have been used also in variants of the covering algorithm implemented in rule learning approaches such as SLIPPER (Cohen, 1999), RL (Lee, Buchanan, & Aronis, 1998) and DAIRY (Hsu, Soderland, & Etzioni, 1998). A variant of the weighted covering algorithm has been used also in the context of confirmation rule subset selection (Gamberger & Lavrač, 2000), used as a basis for the rule subset selection algorithm RSS described in this paper.

## 5.2 Measures of Interestingness

Various rule evaluation measures and heuristics have been studied for subgroup discovery, aimed at balancing the size of a group (referred to as factor $g$ by Klösgen, 1996) with its distributional unusualness (referred to as factor $p$). The properties of functions that combine these two factors have been extensively studied (the so-called "$p$-$g$-space").

Similarly, the weighted relative accuracy heuristic, defined as $WRAcc(Class \leftarrow Cond) = p(Cond) \cdot (p(Class|Cond) - p(Class))$ and used by Todorovski, Flach, and Lavrač (2000), trades off generality of the rule ($p(Cond)$, i.e., rule coverage) and relative accuracy $p(Class|Cond) - p(Class)$. This heuristic is a reformulation of one of the measures used in EXPLORA.

Besides such 'objective' measures of interestingness, some 'subjective' measure of interestingness of discovered patterns can be taken into the account, such as actionability ('a pattern is interesting if the user can do something with it to his or her advantage') and unexpectedness ("a pattern is interesting to the user if it is surprising to the user") (Silberschatz & Tuzhilin, 1995).

## 5.3 Subgroup Evaluation Measures

Evaluation of induced subgroups in the ROC space (Provost & Fawcett, 2001) shows classifier performance in terms of false alarm or *false positive rate* $FPr = \frac{FP}{TN+FP}$ (plotted on the $X$-axis) that needs to be minimized, and sensitivity or *true positive rate* $TPr = \frac{TP}{TP+FN}$ (plotted on the $Y$-axis) that needs to be maximized. The ROC space is appropriate for measuring the success of subgroup discovery, since subgroups whose $TPr/FPr$ tradeoff is close to the diagonal can be discarded as insignificant. An appropriate approach to evaluating a set of induced subgroups is by using the area under the ROC convex hull defined by subgroups with the best $TPr/FPr$ tradeoff as a quality measure for comparing the success of different learners.

Alternatives to the area under the ROC convex hull computation are other standard evaluation measures used in rule learning, such as predictive accuracy or, in the case of time/efficiency constraints that need to be taken into the account, the tradeoff measures DEA (Keller, Paterson, & Berrer, 2000) and Adjusted Ratio of Ratios (ARR) (Brazdil, Soares, & Pereira, 2001) that combine accuracy and time to assess relative performance.

Optimized accuracy is, however, not the ultimate goal of subgroup discovery. In addition to the area under the ROC convex hull quality measure, other important success measures are rule significance (measuring the distributional unusualness of a subgroup), rule coverage (measuring how large is a discovered subgroup), rule size and size of a rule set (measuring the simplicity and understandability of discovered knowledge). These measures were used to evaluate the results of the CN2-SD subgroup discovery algorithm (Lavrač et al., 2002).

## 5.4 Subgroup Visualization

Data visualization methods have been part of statistics and data analysis research for many years. This research concentrated primarily on plotting one or more independent variables against a dependent variable in support of exploratory data analysis (Tukey, 1977; Lee, Ong, & Quek, 1995; Unwin, 2000).

The visualization of analysis results has, however, gained only recently some attention with the proliferation of data mining (Card, Mackinlay, & Shneidermann, 1999; Fayyad, Grinstein, & Wierse, 2002; Keim & Kriegel, 1996; Simoff, Noirhomme-Fraiture, & Boehlen, 2001). The visualization of analysis results primarily serves four purposes: better illustrate the pattern to the end user, enable the comparison of patterns, increase pattern acceptance, and enable pattern editing and support for "what-if questions". The recent interest in the visualization of analysis results was spawned by the often overwhelming number and complexity of data mining results.

Readers interested in comparing the visualization method proposed in this paper with other subgroups visualization methods can find the visualization of subgroups A1–C1 in the joint work by Gamberger, Lavrač, and Wettschereck (2002).

## 6. Conclusions

This paper presents a novel subgroup discovery algorithm integrated into the end to end knowledge discovery process. The discussion and empirical results point out the importance of effective expert-guided subgroup discovery in the $TP/FP$ space. Its main advantages are the possibility to induce knowledge at different levels of generalization (achieved by tuning the $g$ parameter of the subgroup discovery algorithm) used in the rule quality measure that ensures the induction of high quality rules also in the heuristic subgroup discovery process. The paper argues that expert's involvement in the induction process is necessary for successful actionable knowledge generation.

The proposed expert-guided subgroup discovery process consists of the following steps: problem understanding, data understanding and preparation, subgroup discovery, subgroup subset selection, statistical characterization of subgroups, subgroup visualization, their interpretation and evaluation. The main steps, described in detail in this paper, are subgroup discovery and the selection of a subset of diverse subgroups, followed by the statistical characterization of subgroups that adds supporting factors to the induced subgroup descriptions.

Supporting factors represent redundant information about subgroups, but, in our opinion, their function is extremely important in pattern description, because they help the experts to obtain a more complete characterization and better understanding of subgroups. Moreover, they increase the expert's confidence that the pattern is appropriate for the problem that he is trying to solve. In addition, subgroup visualization helps in understanding the relationships among patterns and gives visual insights into their sensitivity and false alarm rate.

The presented approach to descriptive induction uses expert knowledge at every step. Our intention was not to build a system that will replace experts but rather to provide a methodology that will help experts in the knowledge discovery process. In our view, the possibility of guiding the induction process is an advantage of this approach.

## Acknowledgments

This work was supported by the the Croatian Ministry of Science and Technology, Slovenian Ministry of Education, Science and Sport, and the EU funded project Data Mining and Decision Support for Business Competitiveness: A European Virtual Enterprise (IST-1999-11495). We are grateful to Goran Krstačić and Tomislav Šmuc for their collaboration in the experiments in coronary heart disease risk group detection, to Peter Flach for his collaboration in the analysis of different quality measures, and to Dietrich Wettschereck for providing pointers to the related work in visualization.